\documentclass[10pt,conference,compsocconf]{IEEEtran}

\usepackage{times}
\usepackage{epsfig}
\usepackage{graphicx}
\usepackage{calc}
\usepackage{amsmath}
\usepackage{amssymb}
\usepackage{booktabs}
\usepackage{rotating}
\usepackage{caption}
\usepackage{subcaption}
\usepackage{xcolor}
\usepackage{soul}
\usepackage{threeparttable}
\usepackage{multirow}

\usepackage[
  pagebackref=false,
  breaklinks=true,
  colorlinks,
  bookmarks=false]%
{hyperref}

\usepackage{cleveref}
\usepackage{cite}

\DeclareMathOperator*{\argmax}{arg\,max}

\def\IEEEbibitemsep{0pt plus .5pt}
\IEEEtriggeratref{16}

\begin{document}
\title{\LARGE aUToLights: A Robust Multi-Camera\\ Traffic Light Detection and Tracking System\vspace{-2mm}}

\newcommand{\ts}{\hspace*{0.4em}}
\def\onedot{.,\ }

\def\eg{e.g\onedot} \def\Eg{E.g\onedot}
\def\ie{i.e\onedot} \def\Ie{I.e\onedot}
\def\wrt{w.r.t\onedot} \def\dof{d.o.f\onedot}
\def\etal{et al.\ }

\author{
\IEEEauthorblockN{Sean Wu, Nicole Amenta, Jiachen Zhou, Sandro Papais and Jonathan Kelly}
\IEEEauthorblockA{\normalfont Institute for Aerospace Studies,
University of Toronto, Canada\\
\texttt{\texttt{<first\_name>.<last\_name>@robotics.utias.utoronto.ca}}}
}

% https://tex.stackexchange.com/questions/29381/is-it-normal-for-bibtex-to-replace-similar-author-names-with
\bstctlcite{IEEEexample:BSTcontrol}
\maketitle
\begin{abstract}
% 1/4 page
Following four successful years in the SAE AutoDrive Challenge Series I, the University of Toronto is participating in the Series II competition to develop a Level 4 autonomous passenger vehicle capable of handling various urban driving scenarios by 2025.
Accurate detection of traffic lights and correct identification of their states is essential for safe autonomous operation in cities.
Herein, we describe our recently-redesigned traffic light perception system for autonomous vehicles like the University of Toronto's self-driving car, Artemis.
Similar to most traffic light perception systems, we rely primarily on camera-based object detectors.
We deploy the YOLOv5 detector for bounding box regression and traffic light classification across multiple cameras and fuse the observations.
To improve robustness, we incorporate priors from high-definition semantic maps and perform state filtering using hidden Markov models.
We demonstrate a multi-camera, real time-capable traffic light perception pipeline that handles complex situations including multiple visible intersections, traffic light variations, temporary occlusion, and flashing light states. 
To validate our system, we collected and annotated a varied dataset incorporating flashing states and a range of occlusion types.
Our results show superior performance in challenging real-world scenarios compared to single-frame, single-camera object detection.

\end{abstract}

%----- Section ----
\section{Introduction}
% 3/4 page (1.25 cols)
% Why traffic light perception is important
Autonomous vehicles (AVs) have the potential to provide safe, efficient, and accessible transportation.
According to research, 30 to 40 percent of traffic accidents in Canada and the United States occur near intersections \cite{chen2012analysis}.
Therefore, there is an opportunity for AVs to improve road safety by focusing on intersection handling.
To safely operate autonomously at intersections, AVs require the capability to accurately and reliably detect traffic lights (TLs).

\begin{figure}[t]
    \vspace{2mm}
    \centering
    \setlength{\fboxsep}{0pt}%
	\setlength{\fboxrule}{1pt}%
	\fbox{%
    \includegraphics[width=\columnwidth - 2pt,trim=2pt 2pt 2pt 0pt,clip]{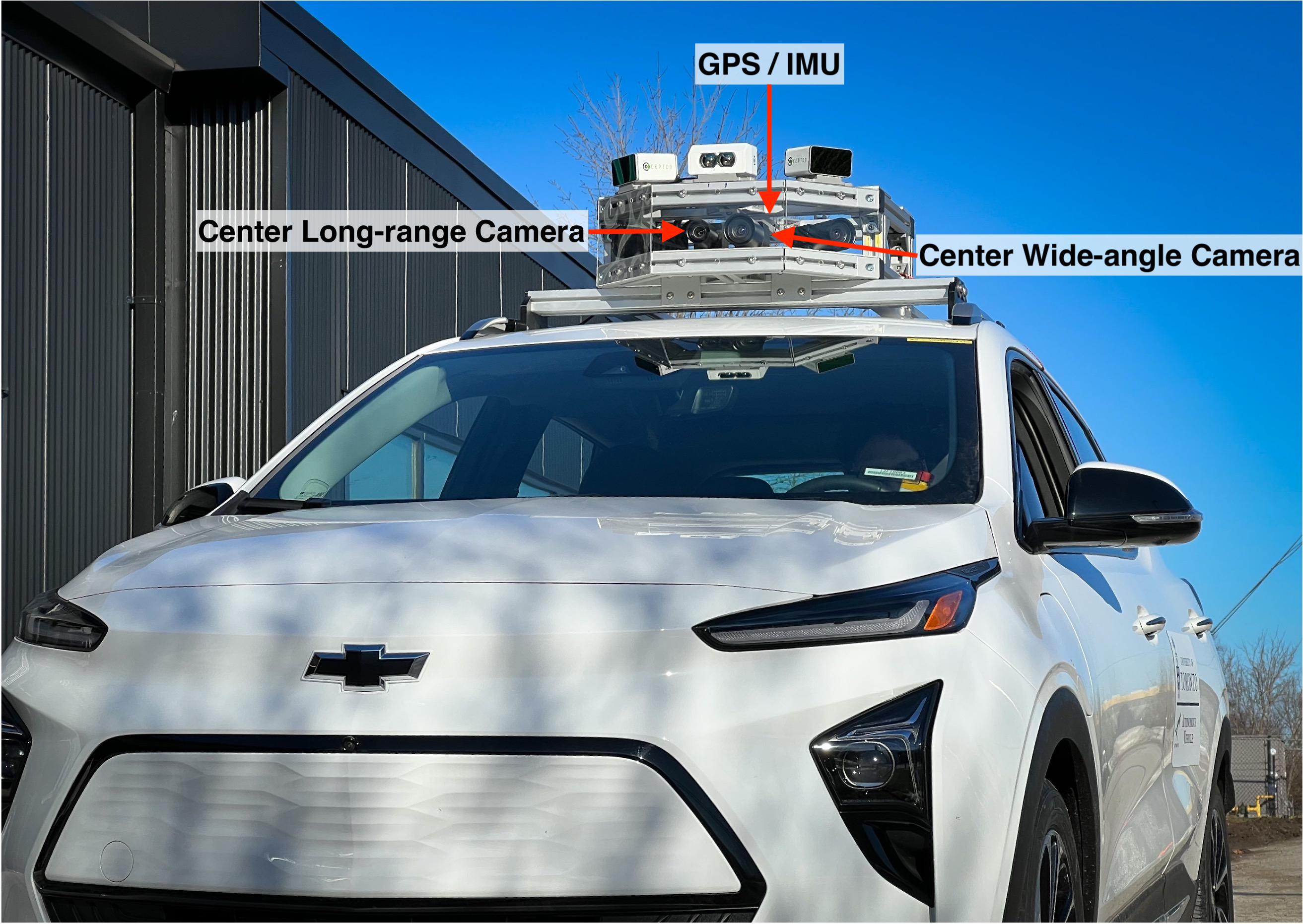}}
    \caption{Our autonomous vehicle, \textit{Artemis}, at the University of Toronto Institute for Aerospace Studies. Two front-facing cameras are used for traffic light detection (one narrow-angle and one wide-angle). A GPS receiver with an integrated IMU is located directly behind the cameras.}
    \label{fig:artemis}
    \vspace{-5mm}
\end{figure}

Camera-based TL detection provides useful semantic information about a given light state.
However, it is necessary to localize TLs in 3D space because, with proper camera coverage, multiple intersections at different distances may be visible within a single image.
Knowledge of the 3D position of each TL makes it easier to determine which light is most relevant to the AV at a given time and to execute the appropriate behavior in the lane leading into the intersection.
Various techniques to infer 3D position include monocular depth estimation, stereo imaging, and LiDAR distance measurement.
However, monocular depth estimation performs poorly in the sky region, stereo imaging is not reliable at long range (when there are only a few pixels on the target), and LiDARs have a limited vertical field-of-view (FOV) that can prevent them from sensing TLs.

In this work, we take an alternative approach to infer the 3D positions of TLs by fusing camera-based detections with high-definition (HD) map data.
Assuming proper sensor calibration and localization, the combined TL detection and HD map approach has the ability to reject false positives while handling false negatives due to occlusion and object detection limitations.
Our previous work \cite{burnett2021zeus,qian2022design} used a map-first approach to TL tracking.
However, HD maps may incorporate errors from a variety of sources, including road construction and maintenance changes.
In this paper, we present a more robust detection-first approach to TL tracking using an improved 2D detector and additional priors.
Our new approach enables us to correct HD map issues, such as missing or extra TLs, by projecting 2D detections to determine 3D positions based on the light class and prior knowledge of the light dimensions.

Most TL perception methods use single-camera, single-frame object detection, while our method instead performs tracking across multiple cameras and across time to ensure temporal consistency.
As shown in \Cref{fig:artemis}, the two front-facing centre cameras allow us to achieve both long-range detection and a wide field of view (FOV).
Multiple cameras also increase accuracy by incorporating more observations of the lights from overlapping views.
Filtering improves robustness and aids temporal consistency, mitigating any erratic behavior in downstream autonomous driving tasks such as planning and control.
Occasionally, single-frame camera-based detectors may produce incorrect predictions or miss detections (i.e., false negatives).
We employ hidden Markov models (HMMs) and leverage the knowledge that TLs are regulated to follow specific transition sequences to improve state estimation.

To evaluate our TL tracking method, we require a dataset that contains challenging real-world lighting scenarios, such as fully occluded or obstructed lights.
We make use of our own TLs to create and annotate a variety of key scenarios, using the programmed light transition sequences in the labelling process to identify fully occluded lights that cannot be seen in a given frame.
Our labelling method is unique from those used for other TL datasets that only include annotations for visible TLs and that do not have access to the light controllers.

% Systems paper contribution: What we seek to do and how we do it
To the best of the authors' knowledge, the systematic integration of real-time object detection, HD map priors, and HMM-based state filtering applied to multi-camera perception has not been previously explored in the literature.
While our TL perception pipeline is optimized for the SAE AutoDrive II competition, our implementation, under competition constraints and with limited resources, could be a valuable reference for other TL perception systems.

%----- Section ----
\section{Related Work}
% 3/4 page

Traffic Light Recognition (TLR) systems must provide accurate position and state information for nearby TLs by detecting and tracking each light.
The detection process involves determining candidate locations and classifying the TL type and signal state.
Detections can then be filtered to yield more consistent results by identifying recurring TL candidates within a sequence of frames.

% Detection
TL detection involves applying classical or deep learning-based computer vision algorithms to camera images.
The output is typically a bounding box that is defined by a centroid position $(c_x, c_y)$, height $h$, and width $w$.
Classical computer vision methods predominantly use color segmentation to detect TLs based on the illuminated bulbs \cite{fairfield2011traffic,jang2014multiple,diaz2015robust}.
Other approaches include template matching \cite{levinson2011traffic} or the use of hand-crafted feature detectors to find TLs based on the shape of the housing or bulbs \cite{nienhuser2010visual,trehard2014tracking}.
However, classical approaches are susceptible to overfitting and may not be as robust to color disturbances, partial occlusion, oblique viewing angles, and false positives from brake lights, spurious reflections, or pedestrian crossing signals \cite{jensen2016vision}.

% Learning approaches
Recent vision-based deep learning approaches such as \cite{ouyang2019deep,wang2018traffic,behrendt2017deep,jensen2017evaluating} have outperformed classical approaches.
While learning methods can achieve better performance, they require more powerful and expensive hardware to operate in real time.
Consequently, detection network model size and inference speed are crucial considerations when choosing the detection method.
For instance, some methods combine a heuristic region of interest (ROI) detector with a lightweight convolutional neural network (CNN) model to achieve real-time performance \cite{ouyang2019deep}.
\begin{figure*}[th!]
    \includegraphics[width=17cm]{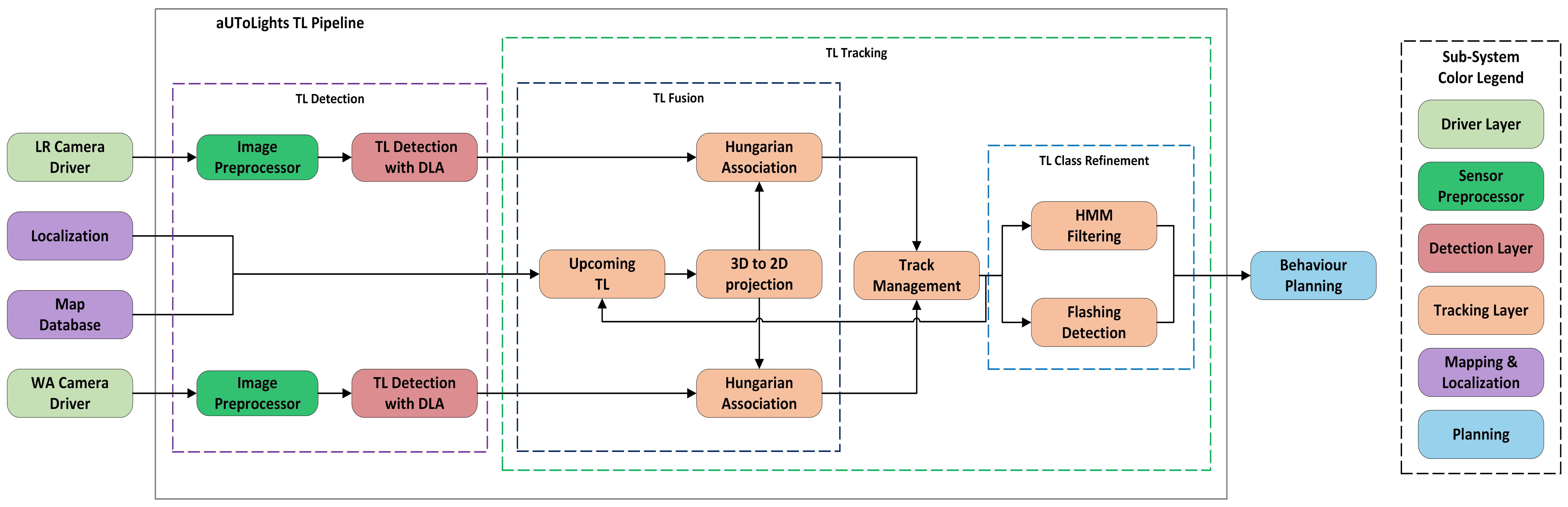}
    \centering
    \caption{The aUToLights traffic light perception pipeline architecture is comprised of TL detection, TL fusion, and TL class refinement, as indicated by the series of dashed boxes.}
    \label{fig:tl_arch}
     \vspace{-3mm}
\end{figure*}

% Maps
A significant improvement in the accuracy of TL detection, regardless of detector type, can be achieved by using HD maps.
HD maps have been applied as priors to improve TL detection and recognition.
One common approach is to use HD maps to generate ROIs to run a TL detector on \cite{fairfield2011traffic,levinson2011traffic,jang2017traffic}.
The ROI is determined by projecting the 3D TL position from the map into the image plane.
Using ROIs can reduce false positives and improve performance by only running the TL detectors when necessary.
ROI generation accuracy heavily depends on the accuracy of camera extrinsic and intrinsic calibration, vehicle localization, and the TL positions encoded within the HD map.
Typically, the ROI is inflated to handle these sources of uncertainty.
Motion compensation is also used by \cite{fairfield2011traffic} and \cite{levinson2011traffic} to improve ROI generation.
However, these ROI-based detection methods are not robust to large errors in the map locations of TLs, localization errors, or new TLs not found in the map.

% State estimation
Once the light position has been identified, the next step is state estimation.
Classical computer vision methods perform TL state classification separately from detection, by extracting features like color from the previously-detected TL candidates \cite{diaz2015robust,fairfield2011traffic} or by using template matching \cite{trehard2014tracking,de2009traffic,de2009real}.
More recently, deep learning models perform detection and classification at the same time to generate the bounding box parameters and class outputs.
For real-time applications, these models are single-stage detection networks that perform regression of bounding boxes and classification in one pass without a second refinement stage.

An early method for single-stage object detection is You Only Look Once (YOLO) \cite{redmon2016you} which follows the novel paradigm of applying a single CNN-based classifier to the whole input image. The network divides the image into a grid of cells and detects objects within each cell using predefined anchor boxes.
The single shot multi-box detector (SSD) \cite{liu2016ssd} improves this approach by using an image feature pyramid to generate features and bounding boxes at multiple scales and hard-negative mining in the loss function.
The RetinaNet method \cite{lin2017focal} introduced the focal loss to address foreground-background class imbalance by assigning more weight to hard misclassified examples and also adding skip-connections to the feature pyramid network.
More recent iterations of YOLO have built on these methods with additional improvements and have become popular for real-time TLR \cite{behrendt2017deep,jensen2017evaluating}

% TL tracking
The TL tracker can be used to improve performance by using correspondences between current bounding box proposals to past proposals and HD map TLs.
This association of TL candidates identifies if a candidate box with the same properties was found in a nearby area and creates a history of detection observations.
Using the class history of a tracked TL, the tracker is able to filter the observations to provide a refined localization and class estimate, reducing errors.
TL tracking can handle false positives and occasional false negatives due to effects like occlusion.
A common approach to handle false positives is to use a track `trial period' to only output new detections if they have been seen for multiple consecutive frames \cite{haltakov2015semantic}.

% Class refinement
Simple tracking methods for class filtering include frequency-based approaches that use the most frequent TL class over a window of $N$ recent frames \cite{diaz2012suspended}.
While this mode filter approach handles occasional 2D misclassifications, it introduces significant latency during transitions between signal states.
For instance, if a TL transitions from red to green, the mode filter will continually output a red state until $\frac{N}{2}$ green frames have been observed.
More advanced tracking methods use Kalman filters \cite{diaz2015robust} and HMMs \cite{nienhuser2010visual,gomez2014traffic} to provide better localization and class estimation.
However, these tracking methods have typically been coupled with classical 2D detectors.
Deep learning methods provide detection confidence values that can be incorporated into these probabilistic tracking methods.
Moreover, these methods have not previously been combined with HD map priors.

% Traffic Light Datasets
Current publicly available TL datasets have significant limitations.
We were unable to find an annotated dataset containing multiple cameras, GPS data, camera intrinsics and extrinsics parameters, and an associated HD map.
Furthermore, TLs are region-specific and there were no annotated datasets with the TL variants that we required.

%----- Section ----
\section{Methodology}
\label{sec:methodology}
% 1-1.5 pages
% State Estimation + Persistence through time
%
Our aUToLights TLR system architecture is shown in \Cref{fig:tl_arch}.
The pipeline has two main stages: TL detection and TL tracking.
In the following sections, we discuss the pipeline stages in detail.

\subsection{Traffic Light Detection}

The TL detection model receives images from multiple cameras.
An image preprocessor module resizes the image to be compatible with the detection input model size.
The detection model is responsible for outputting bounding box candidates and classes for each image from the cameras.
To select a model architecture for TL detection, we previously benchmarked several candidate architectures on the nuImages \cite{nuscenes} object detection benchmark task \cite{qian2022design}.
From benchmarking, YOLOv5 \cite{yolov5} stood out because of its high performance and ability to meet our latency requirements.
An added benefit of the YOLOv5 architecture is the availability of various model sizes: nano, small, medium, large, and extra large.
These models can easily be interchanged for faster testing to meet changing latency requirements.
The YOLOv5 model has been further improved for real-time performance via deep learning acceleration (DLA), which converts the model to a TensorRT format that can be accelerated by hardware-specific optimization.

\subsection{Traffic Light Fusion}
The first half of the TL tracking module is responsible for the association and fusion of TL observations across time.
We associate high-confidence 2D TL detection candidates with the 3D positions of known TLs projected into the image.
The association of lights allows us to combine accurate class predictions inferred from images with precise 3D TL positions from our HD map.
We perform TL association separately for each camera in three steps.

\begin{enumerate}
    \item Determine the set of nearby TLs from the HD map.
    % \item Use TL height and width priors to inflate the 3D centroid $(x,y,z)$ of each upcoming TL into a 3D bounding box.
    \item Project the 3D bounding box of each upcoming TL into each camera sensor's image plane to obtain a projected 2D bounding box $(c_{x,m},c_{y,m},h_m,w_m)$.
    \item Determine the optimal association of each projected TL $(c_{x,m},c_{y,m},h_m,w_m)$ from the map to 2D TL detection candidate bounding boxes $(c_{x,d},c_{y,d},h_d,w_d)$.
\end{enumerate}

% Upcoming TLs
When we initialize the TLR pipeline, we store all TLs from the HD map in a space-partitioning data structure like an R-tree \cite{guttman1984rtree}.
This approach enables us to quickly identify potentially visible TLs by performing efficient spatial range queries using the current vehicle pose and maximum camera sensor range.
The set of potentially visible TLs is later filtered to only include TLs within the field of view based on sensor coverage.

% TL projection
To project the visible TLs, we apply a series of transformations $\mathbf{T}_{BA}$, using 4$\times$4 homogenous transformation matrices, from reference frame $A$ to $B$.
The visible TLs are projected from the Universal Transverse Mercator (UTM) zone frame to the image plane using the time-varying transformation $\mathbf{T}_{\text{cam}\_\text{utm}}(t)$.
Using the exact timestamp $t$ for each camera frame and the localization solution from the GPS/IMU module, we perform pose interpolation to obtain the vehicle pose $\mathbf{T}_{\text{utm}\_\text{ins}}(t)$ at the time when the last image was captured.
Given precise camera calibration, we compute the extrinsic matrix $\mathbf{T}_{\text{cam}\_\text{ins}}$ as
\begin{equation}
    \mathbf{T}_{\text{cam}\_\text{utm}}(t) = \mathbf{T}_{\text{cam}\_\text{ins}} \mathbf{T}_{\text{utm}\_\text{ins}}^{-1}(t).
    \label{eq:T_cam_utm}
\end{equation}
We use $\mathbf{T}_{\text{cam}\_\text{utm}}(t)$ to transform the 3D TL bounding box coordinates, denoted as $\mathbf p_{\text{cam}}$, to be relative to the camera frame. Next, each point $\mathbf p_{\text{cam}}$ is transformed into pixel coordinates, $(u,v)$,
\begin{equation}
    \label{eq:projection}
    \begin{bmatrix}
        u \\
        v \\
        1
    \end{bmatrix}
    =
    \mathbf{K}\,\mathbf p_{\text{cam}} \frac{1}{z_{\text{cam}}},
\end{equation}
where $\mathbf{K}$ is the camera intrinsic calibration matrix.

\begin{figure}[t!]
    \centering
    \setlength{\fboxsep}{0pt}%
	\setlength{\fboxrule}{1pt}%
	\fbox{%
    \includegraphics[width=\columnwidth - 2pt]{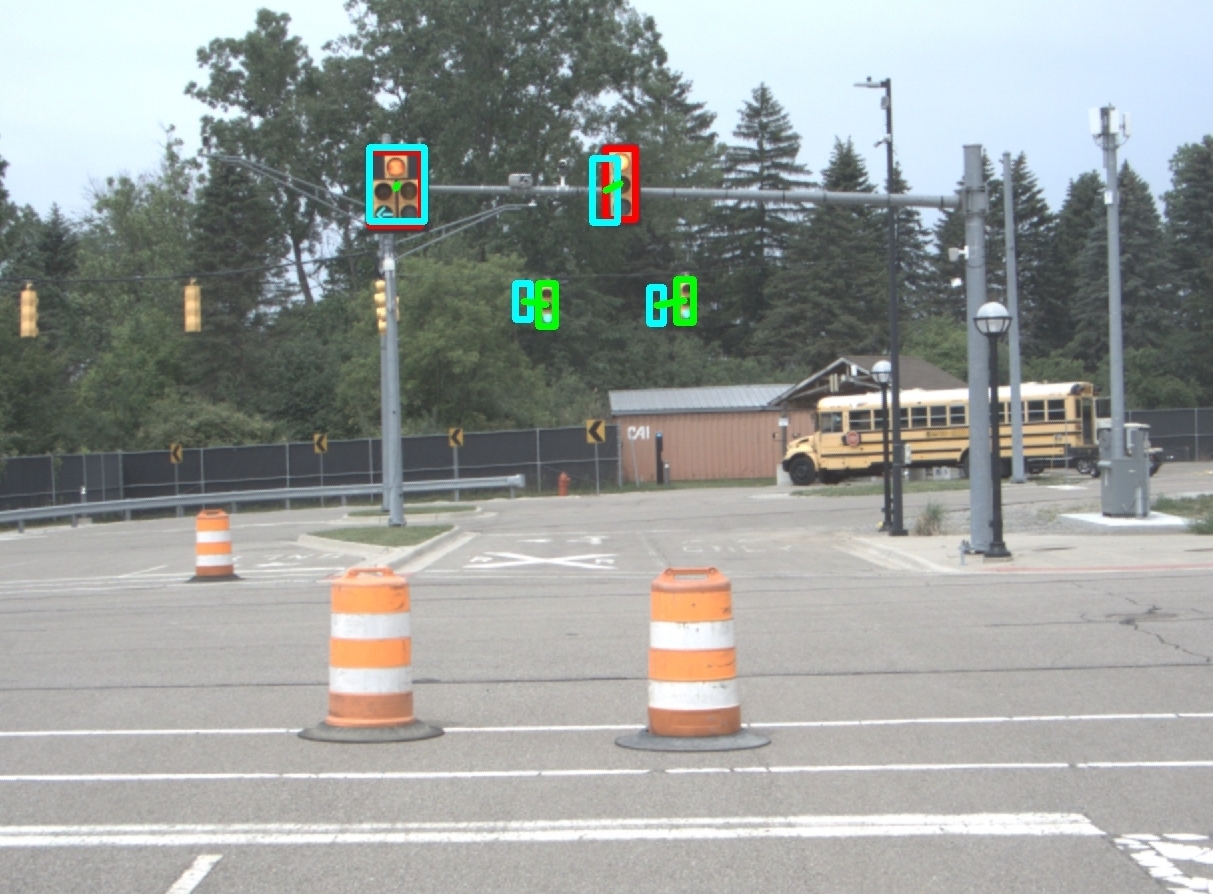}}
    \caption{An example of TL association with two visible intersections. Red and green boxes represent 2D TL detections and their states, indicating red and green lights respectively. Cyan boxes indicate TLs projected from the HD map. Associations are depicted by green lines.}
    \label{fig:association}
    \vspace{-3mm}
\end{figure}

% TL association
After projecting the TLs from the map onto the image plane, each TL needs to be associated with a detected TL bounding box.
Consider $M$ projected TLs $(c_{x,m},c_{y,m},h_m,w_m)$ from the map and $N$ detected TL bounding boxes $(c_{x,d},c_{y,d},h_d,w_d)$.
We must find the optimal one-to-one pairing of projected map TLs and detected TL bounding boxes.
Each pairwise association of the $i$-th projected TL and $j$-th bounding box is assigned a cost $c(i,j)$ with the following metrics.
\begin{enumerate}
    \item \textbf{L2 distance} between the projected TL bounding box and detected 2D bounding box. A maximum distance threshold is applied, where the cost $c_{L2}$ is set to $\infty$ if the threshold is exceeded.
    $c_{L2} = ||(c_{x,m},c_{y,m},h_m,w_m)-(c_{x,d},c_{y,d},h_d,w_d)||_2$.
    \item \textbf{Type difference} between the TL type in the map and the 2D detected class. If the types do not match, the cost is set to $\infty$. Otherwise, it is set to 0.
\end{enumerate}
\begin{equation}
    c(i,j) = c_{L2} + c_{\text{type}}.
    \label{eq:cost}
\end{equation}
Using these metrics, we construct an $M \times N$ matrix of costs.
For the $i$-th projected TL, we want to find the set of assignments $\{(i,j)\}$ corresponding to minimum total cost $\sum_{\{(i,j)\}} c(i,j)$.
The set cannot repeat any $i$ or $j$ in multiple pairs.
Infinite costs $c(i,j)=\infty$ are considered invalid assignments and are ignored.
We use the Hungarian association algorithm to find the optimal set of assignments in $O(n^3)$ time \cite{huang2008robust}.
After association, the 3D position for each 2D detection can simply be read from the map for the associated projected TL.
An example of the association process is shown in \Cref{fig:association}.
The figure shows a scenario where 3D TL positions are required to determine which TLs belong to the nearest upcoming intersection when multiple TLs are visible in the distance.

Our pipeline processes the candidate bounding boxes using the HD map to improve the localization of lights.
The TLR system should be robust to mapping errors in the case that a light is not present in the HD map or an incorrect light exists in the map due to light installation, removal, or a mapping oversight.
In the case of an incorrect light, there will be a light in the map without any associated detections, so the light will not be tracked.
In the case of a missing light, the detector predicts a 2D bounding box that will not be matched to any known TL in the map; we instead project the 2D detection to 3D space for tracking.
To compute this projection, we leverage the standard physical size of TLs as a reference to derive the depth position from the detected 2D bounding box dimensions.
The TL 3D centroid location is then calculated by projecting the 2D bounding box center $[c_x, c_y]$ to 3D using the estimated depth.

\subsection{Track Management}
% track trial
The goal of track management is to fuse shared observations of the same light across cameras and time.
A track is created for each unique traffic light that is observed and each tracked light is assigned a track ID to uniquely identify it.
A TL from the map must be detected for $N_\text{birth,map}$ consecutive frames before a TL track using the 3D map position is created.
Any 2D detections that have been identified for $N_\text{birth,2D}$ consecutive frames and have not been associated are used to create a TL track with the projected 3D position.
We assume any false positives generated by the 2D detector are
uniformly distributed (sparse).
Since false positives are unlikely to consistently appear in the same location, they will not spawn TL tracks because tracks are only created after $N_\text{birth,2D}$ successive observations.
%

% Track update
For each frame $f$, the TL track history is updated by fusing the 3D position from the map and the 2D detected class into the latest observation $O_f = (c_{x,m},c_{y,m},h_m,w_m,\text{type})$.
For each tracked TL, we maintain a history of the last $N$ observations $\mathbf{O} = O_1, O_2, \dots, O_N$.
Detections of a TL track from different cameras are treated as separate observations.

% track deletion
Any TLs that are not observed for more than $N_\text{death}$ are not reported.
For TL tracks that are not in the map, we keep their 3D position stored in our spatial data structure.
Since TLs are static objects, we are confident in the TL's position, but we discard the predicted state after $N_\text{death}$ frames.
This tracking approach allows us to handle 2D false negatives due to temporary occlusions and recover from inaccurate HD map information.
The $N_\text{birth,map}, N_\text{birth,2D}, N_\text{death}$ parameters can be tuned, but we use $N_\text{birth,map} = N_\text{birth,2D} = 2$ and $N_\text{death} = 15$.

\subsection{Traffic Light Class Refinement}
% State Estimation + Persistence through time
For each incoming observation, we refine the initial detected 2D class and smooth detections temporally.
Since TLs have regulated sequences, we can use HMMs \cite{murphy2012machine,rabiner1988tutorial} to estimate the ground truth state by incorporating TL state transition probabilities with our 2D detector class confidence values.
The HMM hidden state at time $t$ is the ground truth TL state $q_t$, and the possible states are $\mathbf{S} = \{S_1, S_2, \dots S_N\}$.
The HMM observed state at time $t$ is the  detected 2D state $o_t$, and the possible states are $\mathbf{V} = \{V_1, V_2, \dots V_N\}$.
We represent the HMM observed and hidden states using the TL states, where $\mathbf{S} = \mathbf{V} = \{\text{3-red}, \text{3-green}, \dots, \text{4-rleft}, \text{4-gleft}, \dots, \text{5dh-red}, \text{5dh-green}\}$.
The HMM transition matrix $\mathbf{A} = \{a_{ij}\}$ and observation matrix $\mathbf{B} = \{b_{ij}\}$ are
\begin{equation}
    \label{eq:hmm_transition_matrix}
    a_{ij} = P(q_t=S_j | q_{t-1}=S_i) \quad 1 \leq i, j \leq N,
\end{equation}
\begin{equation}
    \label{eq:hmm_observation_matrix}
    b_{ij} = P(o_t=V_j | q_t=S_i) \quad 1 \leq i, j \leq N.
\end{equation}
Using the object detector confusion matrix and experimental tuning, we determine the state transition matrix, $\mathbf{A}$, observation probability matrix $\mathbf{B}$, and initial state matrix $\mathbf{\pi}$ for each HMM.
To encode the valid TL sequences in $\mathbf{A}$ for each TL type, we use the appropriate subset of valid states in $\mathbf{V}$ and $\mathbf{S}$ to
create a separate HMM $\lambda_i=\{\mathbf{A_i, B_i, \pi_i}\}$, for each type of traffic light.
We use the forward algorithm to efficiently compute the filtered belief state $\mathbf{\alpha_t}$ at each timestep $t$ \cite{murphy2012machine}.
We maintain the belief state $\mathbf{\alpha_t}$ that denotes the probability of the TL being in each state in $\mathbf{S}$ given the past observation sequence $o_{1:t}$.
The predicted true state is simply given by $q_t = \argmax_j \alpha_t(j)$,
\begin{equation}
    \label{eq:hmm_belief_state}
    \alpha_t(j) \equiv P(q_t=S_j|o_{1:t}) \quad 1 \leq j \leq N.
\end{equation}
The HMM filter updates the belief state using the prediction step,
\begin{equation}
    \label{eq:hmm_prediction_step}
    \mathbf{\alpha_t} \propto \mathbf{c_t} \odot (\mathbf{A^T} \mathbf{\alpha_{t-l}}),
\end{equation}
where $\odot$ indicates element-wise multiplication.
We compute the local evidence vector $\mathbf{c_t}$ using Bayes rule with the 2D detector output class confidence vector $\mathbf{x}$ where $x(k) = P(o_t = V_k)$,
\begin{align}
    \label{eq:hmm_evidence_vector_bayes}
    c_t(j) &= P(o_t|q_t=S_j) \quad 1 \leq j \leq N \\[2mm]
    c_t(j) &= \sum_k \frac{P(o_t=V_k) P(q_t=S_j | o_t=V_k)}{P(q_t=S_j)}
\end{align}
To simplify the evidence vector $\mathbf{c_t}$ computation, we assume a uniform distribution for our prior $P(q_t=S_j)$ which is normalized out in the belief state update in \Cref{eq:hmm_prediction_step} to obtain
\begin{equation}
    \label{eq:hmm_evidence_vector_simplified}
    \mathbf{c_t} \propto \mathbf{C}\mathbf{x}.
\end{equation}
The matrix $\mathbf{C}$, where $C(j,k) = P(q_t=S_j | o_t=V_k)$, is computed from the 2D detector confusion matrix.

For flashing light detection, we use a simple duty cycle threshold logic on the unfiltered past observations $\mathbf{O}$.
If the ratio of on and off TL states are both within the duty cycle threshold, then the TL is predicted to be flashing.
Based on the US federal highway regulations on flashing operation of traffic control signals, we use a minimum and maximum duty cycle threshold of $(\frac{1}{2}, \frac{2}{3})$ \cite{fhwa2009mutcd}.

%----- Section ----
\section{Data Collection and Labelling Strategy}

We collected a dataset for training our TL detection model and validating the entire TL perception pipeline\footnote{\url{https://www.autodrive.utoronto.ca/datasets-autolights}}.
The dataset was collected at the University of Toronto Institute for Aerospace Studies (UTIAS) and the University of Michigan MCity Test Facility (MCity).
Our dataset contains variety in TL distances, viewing angles, daytime lighting conditions, and weather.

The MCity portion of the dataset was collected at the 2022 AutoDrive II competition and contained full-sized intersections at distances of 5 metres to 80 metres.
The UTIAS portion of the dataset was collected on a closed track with traffic lights arranged in a mock intersection at distances of 30 metres to 50 metres. The TL states we included belong to a subset of standard Michigan and Ontario traffic signals.
Table \ref{tab:dataset_breakdown} provides a class breakdown of the data we collected, as well as the corresponding training and test set split. The proportion of labels in the test set is higher than in the training set because the training set samples form a subset of the images collected.

\begin{table}[t!]
\caption{Number of training and test samples across varying classes in our dataset. }
\begin{tabular*}{\linewidth}{@{\extracolsep{\fill}}lcccc}
\hline
\textbf{Class} & \textbf{Total} & \textbf{Train} & \textbf{Test} \\ \hline
3-green & 5,919   & 2,413 & 3,507 \\
3-red & 12,296  & 3,791 & 8,505 \\
3-yellow & 1,773 & 773& 1001\\

4-gleft & 3,064 & 1,167& 1,897\\
4-off & 806 & 806 & 0 \\
4-rleft & 3,534 & 1,236& 2,298\\
4-yleft1& 719 & 419& 300\\
4-yleft2& 3299 & 1073& 2226\\

5dh-green & 2,269 & 1,589 & 680\\
5dh-red & 5,713 & 3,444& 2,269\\
5dh-red-gleft & 4,754 & 2,464& 2,290\\
5dh-red-yleft & 1,632  & 1,366 & 266\\
5dh-yellow & 1,772 & 1,486 & 286 \\ \hline
\end{tabular*}
\label{tab:dataset_breakdown}
\vspace{-3mm}
\end{table}

For training samples, we only labelled TLs with visible light states that the detection model can identify.
For test samples, we labelled obstructed and fully occluded TLs since the tracking pipeline should be able to correctly identify these TLs.
Our dataset contains the Michigan flashing yellow arrow signal, which alternates between an off state (all light bulbs turned off) and an on state (only the 4-yleft2 light bulb is on).
For our training data, we annotated the off state as 4-off and the on state as 4-yleft2.
For our test data, we annotated both the on and off states during flashing as a single 4-yleft2 state because the tracking pipeline should indicate that the light is continuously flashing.
Note that during normal operation, the 4-off state is only present during flashing, so we train the detector to detect the 4-off state, but there are no 4-off labels in our test dataset.
This approach was taken in an effort to properly train the TL detection model and evaluate the TL tracker's ability to leverage temporal information to detect flashing and address temporary occlusions.

%----- Section ----
\section{Experimental Results}
Our system was evaluated using Artemis, an AV test platform at the University of Toronto. The sensors used include an inertial navigation system (GNSS/IMU) for localization and the two front-facing cameras as shown in \Cref{fig:artemis}. The center long-range camera has a maximum effective range for TL detection of 64 metres with a $47.3^{\circ}$ horizontal FOV, and the center wide-angle camera has a maximum effective range of 30 metres with an $85.7^{\circ}$ horizontal FOV.

The YOLOv5 \cite{yolov5} object detection model was trained on a server with two NVIDIA RTX3090 GPUs with a cyclic cosine annealing learning rate schedule with a max learning rate of 0.008. The model was pre-trained on the COCO dataset \cite{cocodataset} and fine-tuned for 20 epochs with a batch size of 16 on our own TL dataset. Several common data augmentations were used such as gaussian blur, color jitter, rotations, and flips.

Our method was evaluated on our own test dataset, consisting of around 10,000 images spread across the two data collection locations: Toronto, Ontario, and Ann Arbor, Michigan. The models were evaluated on a server with a single RTX3090 GPU.
The object detection method using YOLOv5 was evaluated for three different model sizes as is shown in \Cref{tab:res_2d}.
The primary evaluation metrics for the object detector include Average Precision (AP) and frames per second (FPS). We report AP averaged over all IOU thresholds, AP at IOU thresholds 0.5 (AP50) and 0.75 (AP75). We also report AP for small (AP$_S$), and medium (AP$_M$) objects with area $<$ (32px)$^2$ and (32px)$^2$ $<$ area $<$ (96px)$^2$, respectively. Our analysis shows that the model performance is similar across model sizes for our dataset.

\begin{table}[t!]
\caption{Object detection evaluation results on our test dataset for three different model variants.}
\begin{tabular*}{\linewidth}{@{\extracolsep{\fill}}ccccccc}
\hline  & FPS & mAP & AP$_{50}$ & AP$_{75}$ & AP$_S$ & AP$_M$ \\
\hline YOLOv5n & 163 & 0.77 &  0.93 & 0.89 &  0.74 &  0.79  \\
YOLOv5s & 109 & 0.77 & 0.93 & 0.86 & 0.72 & 0.82  \\
YOLOv5m & 54 & 0.75 & 0.92 & 0.86 & 0.71 & 0.79  \\
\hline
\end{tabular*}
\label{tab:res_2d}
\end{table}

\begin{table}[t!]
\centering
\caption{Pipeline evaluation results on our test dataset for ablation of various system software components.}
\begin{tabular}{cccccc}
\hline  & FPS $\uparrow$ & APE (m) $\downarrow$ & Class Accuracy $\uparrow$ \\
\hline OD & 109 & 1.19 & 92.8 \\
OD + Fusion & 57 & 0.39 & 92.8  \\
OD + Fusion + Tracking & 54 & 0.39 & 96.0  \\
(Proposed) & & &  \\
\hline
\end{tabular}
\label{tab:res_3d}
\vspace{-0.5cm}
\end{table}

\begin{figure}[b!]
    \centering
    \includegraphics[width=0.48\textwidth]{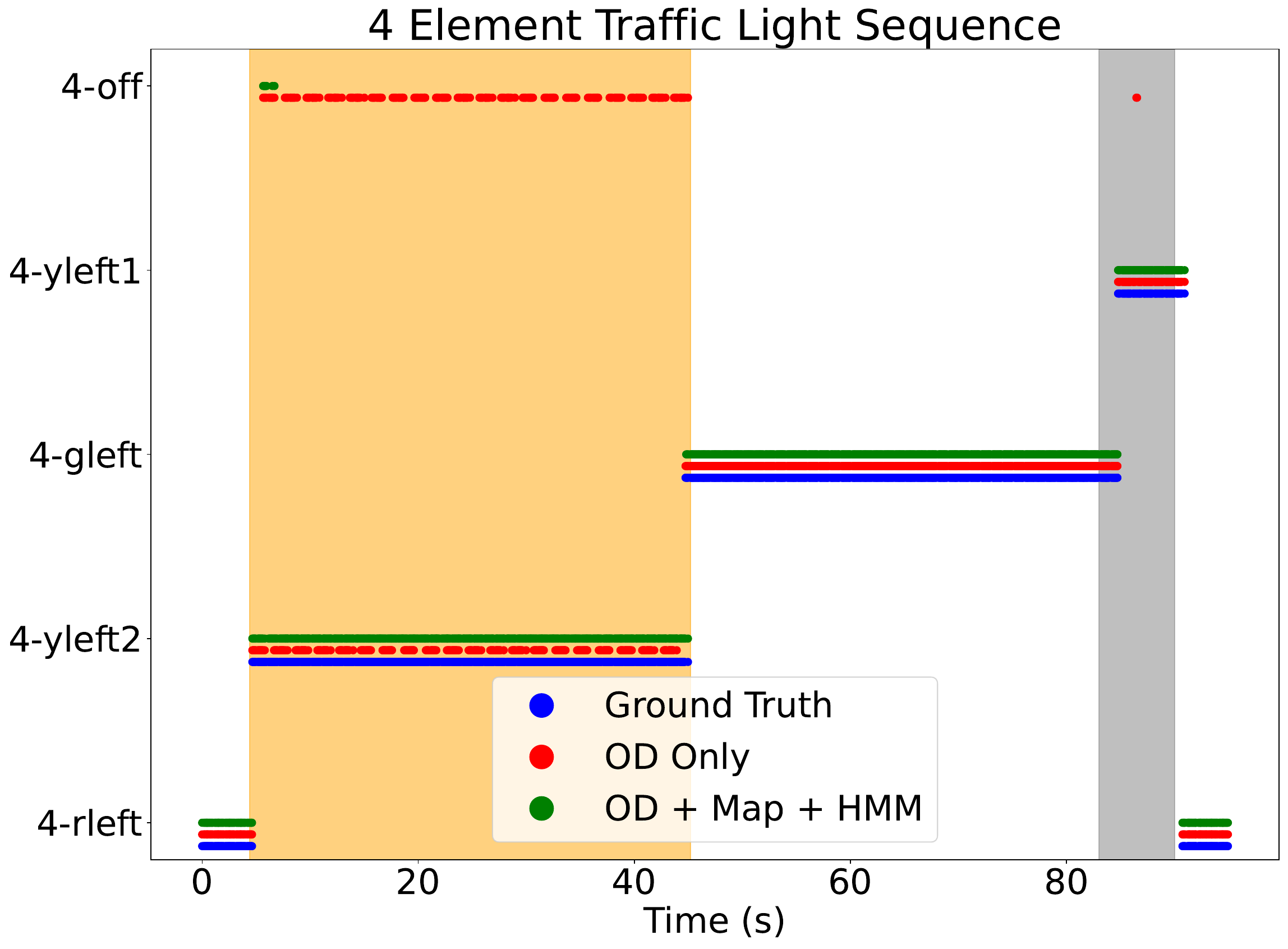}
    \caption{Visualization of a traffic light sequence. The orange region indicates a flashing light and the grey region indicates TL occlusion by a pedestrian. Without tracking, the OD only approach is unable to handle these scenarios.}
    \label{fig:occlusion_sequence}
\end{figure}

The aUToLights pipeline is implemented using C++ and Robotic Operating System 2 (ROS 2) \cite{macenski2022robot}.
We evaluated the pipeline in terms of end-to-end performance and conducted an ablation study for three different variants of our approach, as shown in \Cref{tab:res_3d}.
Evaluation metrics include classification accuracy along with the average position error (APE) in 3D space (i.e., the Euclidean distance between ground truth and predicted light locations).
The first method uses object detection (OD) alone to infer the 3D bounding boxes and classes of the TLs. The second method adds the use of TL Fusion and the HD map as a prior to help with localization. The third method, which we propose, adds track management and class refinement to improve classification performance for occlusions, detection errors, and flashing lights. We show that our proposed approach not only achieves the best APE, it significantly increases classification accuracy. At 54 FPS, real-time performance can be achieved.

The performance of the TL pipeline on a sequence with flashing lights and occlusion is shown in \Cref{fig:occlusion_sequence}. This sequence shows the stability of the output TL predictions and its response latency to changes in the TL state. This also demonstrates the performance improvements from the standalone detection approaches and tracking method compared to the ground truth.
Note that during the flashing sequence, there is a fixed latency before the proposed method correctly identifies the 4-yleft2 flashing state.
This latency is due to the minimum duty cycle threshold in the flashing light detection logic.

\begin{figure}[b!]
    \centering
    \includegraphics*[width=8.5cm]{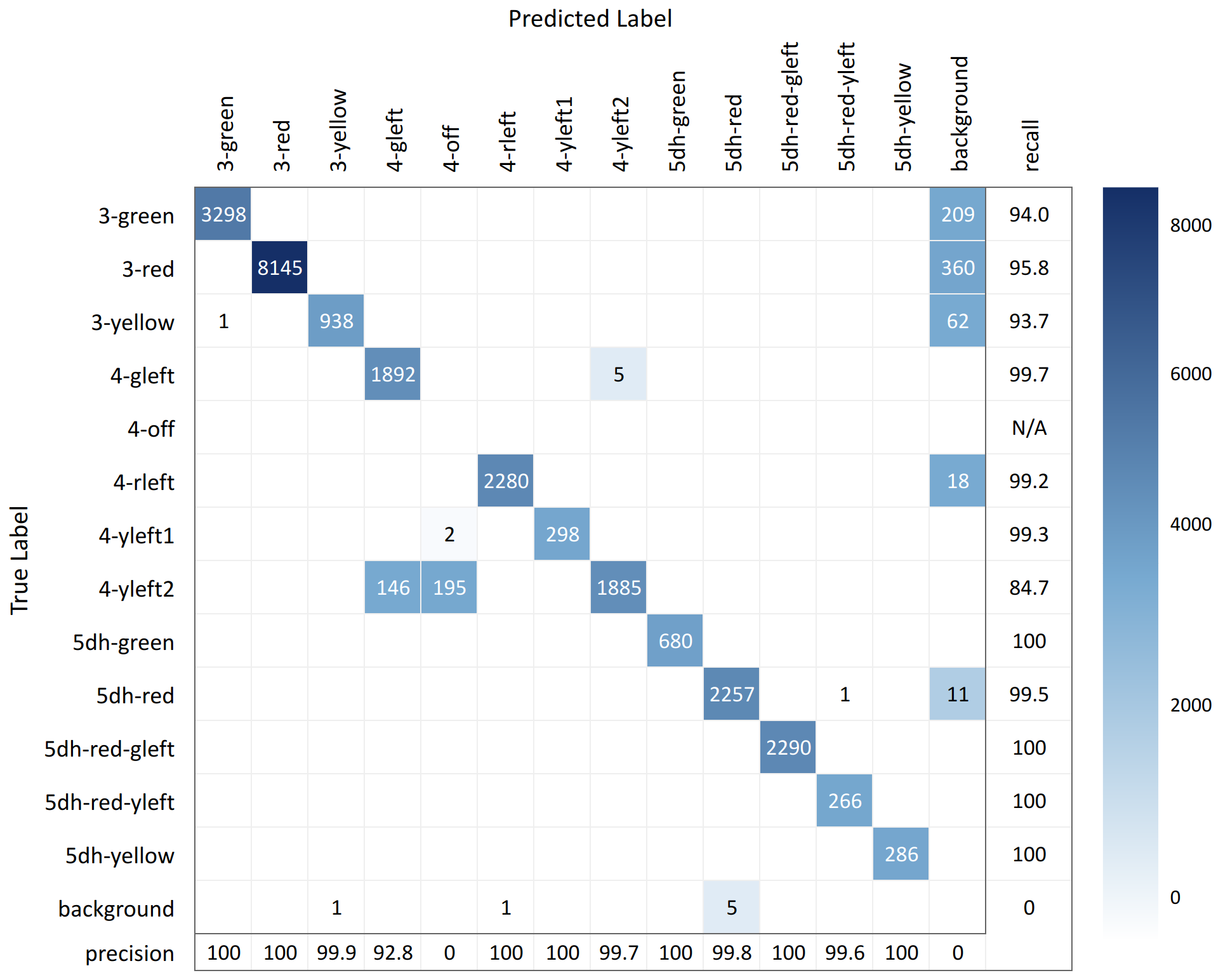}
    \caption{Confusion matrix for our proposed aUToLights TLR pipeline's performance on the aUToLights test dataset.}
    \label{fig:confusion_matrix}
    \vspace{-1mm}
\end{figure}

% confusion matrix
The classification performance of our TL pipeline is demonstrated by the confusion matrix in \Cref{fig:confusion_matrix}.
The confusion matrix shows 659 false negatives (FN) where the predicted label is the background class.
Some of these FNs are due to experiment artifacts where ROS 2 messages were dropped.
Errors due to the detector and tracker (e.g., incorrect association during TL fusion) can also contribute to FNs.
However, some FNs are expected as a tradeoff in our approach because we do not output all detections immediately and instead require $N_\text{birth}$ successive observations to reduce false positives.
The confusion matrix also shows that the 4-yleft flashing state is confused with the 4-gleft2 class and the 4-off class.
The 146 incorrect 4-gleft predictions are caused by a detector error.
The 195 incorrect 4-off predictions are due to latency in the simple flashing light detection approach.

\begin{figure}
    \centering
    \includegraphics[width=0.48\textwidth, trim={0 1cm 0 0}]{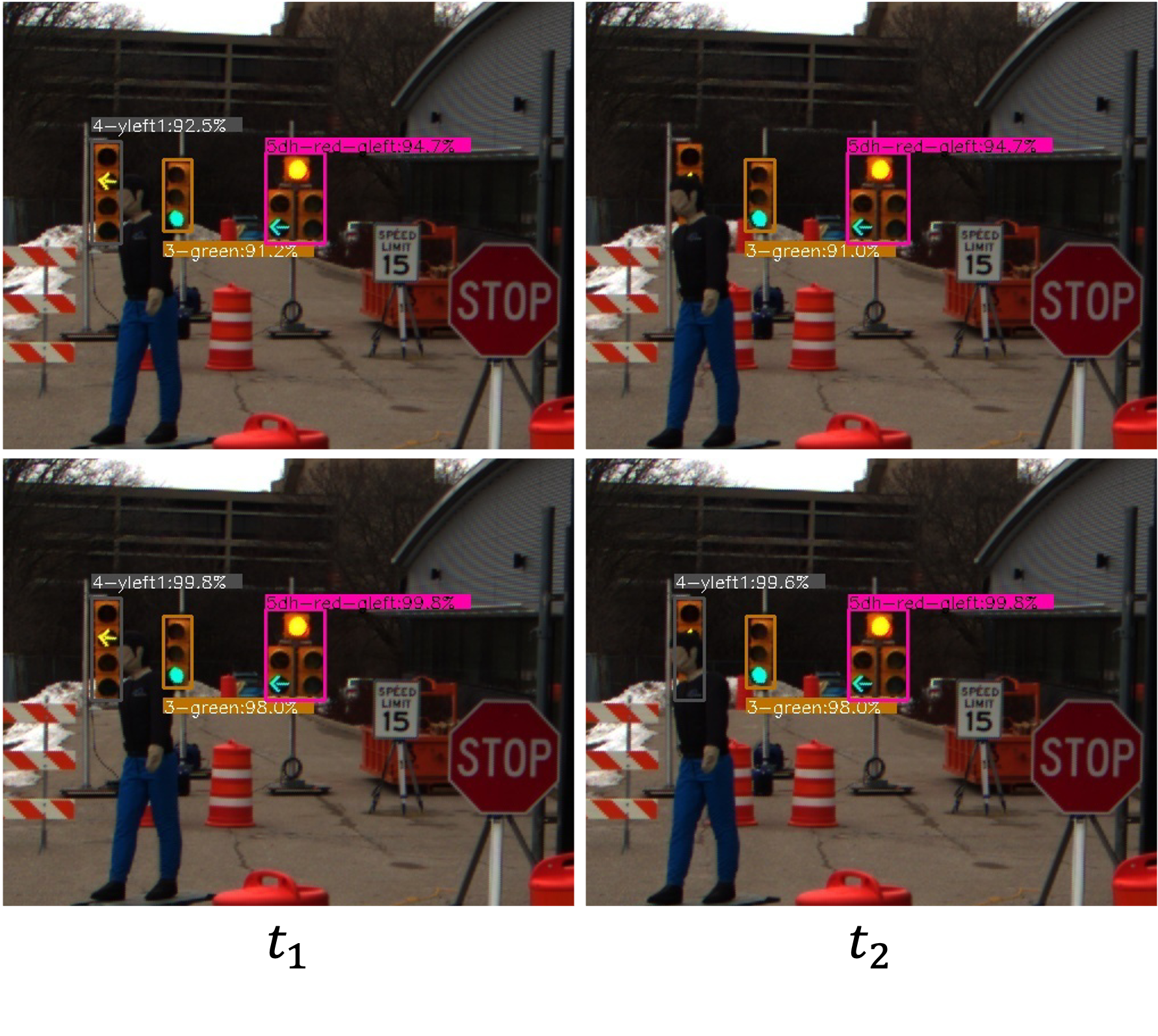}
    \caption{Visualization of TL bounding boxes and class prediction labels across a sequence of frames (examples of two are shown), using OD only approach (\textit{top}) and our proposed tracking method (\textit{bottom}).}
    \label{fig:2DOD_vs_map_hmm_tracking}
    \vspace{-3mm}
\end{figure}

In terms of the overall classification results, the accuracy is 96.0\%.
This is a substantial improvement from the detection-only approach due to the ability to correct detection errors, classify lights with occlusion, and determine transient states such as flashing lights. This improvement is also highlighted in the visualization in \Cref{fig:2DOD_vs_map_hmm_tracking}, indicating that the HMM is able to track through occlusions.

%----- Section ----
\section{Conclusion}

We have presented a novel multi-camera TL detection and tracking system that combines a real-time object detector with HD map priors and HMM state filtering. We have also contributed a labelled TL dataset that captures TLs in different scenes and with varying levels of occlusion. Our end-to-end pipeline provides fast and robust TL detection and can be used to inform vehicle planning. As future work, performance could potentially be enhanced by incorporating lidar measurement updates and Kalman filtering to improve the 3D TL position estimates. Our approach may then have applications for object detection, pushing towards Level 4 autonomy.

\section{Acknowledgments}
This work would not have been possible without the contributions of our sponsors: SAE, General Motors, Lucid Vision Lab, Novatel, REDARC, Ushr, Cepton, and Newark element14. We thank the University of Toronto for providing funding to participate in the SAE AutoDrive Challenge II. Thank you to all members of aUToronto for their contributions, especially Jacob Deery who led driver setup and synchronization for all sensors, and Amy Chen who contributed to dataset collection and annotation.

%%%%%%%% REFERENCES
\bibliographystyle{IEEEtran}
\bibliography{2023-autoronto-lights-crv}

\end{document}